\documentclass{article}
\usepackage{algorithm} 
\usepackage[numbers,sort&compress]{natbib}
\usepackage{algpseudocode} 
\usepackage{PRIMEarxiv}
\usepackage[table,xcdraw]{xcolor}
\usepackage[utf8]{inputenc} 
\usepackage[T1]{fontenc}    
\usepackage{hyperref}       
\usepackage{url}            
\usepackage{booktabs}       
\usepackage{amsfonts}       
\usepackage{nicefrac}       
\usepackage{microtype}      
\usepackage{lipsum}
\usepackage{fancyhdr}       
\usepackage{graphicx}       
\graphicspath{{media/}}     
\usepackage{adjustbox}
\usepackage{algorithm} 
\usepackage{algpseudocode}
\usepackage[section]{placeins}
\usepackage{amsmath}
\usepackage{amssymb}
\usepackage{booktabs}
\usepackage{multirow}
\usepackage{multirow,diagbox,tabularx,blindtext}
\usepackage{threeparttable}
\usepackage{color}
\usepackage{float}
\usepackage{tabularray}
\usepackage{longtable}
\usepackage[numbers,sort&compress]{natbib}

\usepackage{lineno}
\nolinenumbers 
\title{Enhancing Nighttime Vehicle Detection with Day-to-Night Style Transfer and Labeling-Free Augmentation 

}

\author{
 Yunxiang Yang, Hao Zhen, Yongcan Huang, Jidong J. Yang*\\
 Smart Mobility and Infrastructure Lab\\
 College of Engineering\\
 University of Georgia, Athens, GA, USA\\
  \texttt{\{yyang117, Hao.Zhen, yh15555, Jidong.Yang\}@uga.edu} \\
}

\begin{document}
\maketitle

\begin{abstract}
Existing deep learning-based object detection models perform well under daytime conditions but face significant challenges at night, primarily because they are predominantly trained on daytime images. Additionally, training with nighttime images presents another challenge: even human annotators struggle to accurately label objects in low-light conditions. This issue is particularly pronounced in transportation applications, such as detecting vehicles and other objects of interest on rural roads at night, where street lighting is often absent, and headlights may introduce undesirable glare. This study addresses these challenges by introducing a novel framework for labeling-free data augmentation, leveraging CARLA-generated synthetic data for day-to-night image style transfer.  Specifically, the framework incorporates the Efficient Attention Generative Adversarial Network for realistic day-to-night style transfer and uses CARLA-generated synthetic nighttime images to help the model learn vehicle headlight effects. To evaluate the efficacy of the proposed framework, we fine-tuned the YOLO11 model with an augmented dataset specifically curated for rural nighttime environments, achieving significant improvements in nighttime vehicle detection. This novel approach is simple yet effective, offering a scalable solution to enhance AI-based detection systems in low-visibility environments and extend the applicability of object detection models to broader real-world contexts.
\end{abstract}


\section{Introduction}

Accurate and reliable vehicle detection is essential for a wide range of transportation applications, such as traffic monitoring and incident management. However, large performance gaps exist for vehicle detection between daytime versus nighttime conditions. This especially true for rural environments, where streetlighting is often absent.  The disproportionate nighttime fatalities in rural areas compared to urban areas have long been recognized \cite{nhtsa2010}. Nighttime vehicle detection presents unique challenges, including limited visibility, unpredictable lighting conditions, and lower resolution from standard roadside cameras compared with daytime scenarios. In rural settings, where lighting infrastructure is often sparse or non-existent, these challenges are even more pronounced, coupled with the headlight glare that further aggravates the issue \cite{nhtsa_glare}.  Advances in object detection technology must address these challenges to offer reliable performance in such undesirable conditions.

Computer vision techniques leverage appearance information like color, shape, or typical vehicle patterns to detect vehicles from different views achieve good performance \cite{mittal2023ensemblenet, bie2023real}, but most of these works address the problem during the daytime. At night, above appearance features become invalid, and headlights/taillights are almost the only
obvious features. Efforts for nighttime vehicle detection have made significant strides in recent years, particularly in applications that utilize ego cameras and roadside cameras. 

Ego cameras on vehicles offer a "driver" perspective and are primarily utilized in autonomous driving applications. Nighttime vehicle detection using ego cameras has been studied due to the relatively higher quality of data and less challenging nature of the tasks. Their primary focus is on accurately identifying nearby vehicles, a task essential for real-time decision-making and navigation for the safety of the ego vehicle. For ego camera based nighttime vehicle detection task, current researches utilize Generative Adversarial Networks (GANs) and image-to-image translation techniques have applied \cite{huang2018auggan, bang2024semantic,shao2020feature} for enhancing object detection in challenging scenarios such as nighttime and adverse weather conditions. Common approaches focus on translating images from a source domain (e.g., daytime) to a target domain (e.g., nighttime) while preserving critical object features. Models like AugGAN \cite{huang2018auggan} and CycleGAN \cite{zhu2017unpaired,bang2024semantic,shao2020feature} are popular approaches that leverage structure-aware mechanisms to maintain semantic and geometric consistency during style transfer. Techniques such as semantic segmentation and geometric attention maps \cite{bang2024semantic} further ensure that essential object details are retained, enabling robust object detection performance in the target domain. These models generate high-quality synthetic datasets that mimic target domain characteristics, which are then utilized to train and fine-tune object detection models, resulting in improved accuracy and robustness.

In addition to domain translation, cross-domain learning techniques are integrated to bridge the performance gap between source and target domains. For instance, Convolutional Block Attention Mechanisms (CBAM) \cite{xu2023cross} enhance detection accuracy by focusing on salient image regions, while feature enhancement modules, fuse daytime and nighttime data to mitigate ambient light interference. Furthermore, advanced loss functions and data augmentation strategies refine model training, addressing challenges like reduced visibility and occlusion. Collectively, these methodologies highlight the efficacy of GAN-based frameworks, feature enhancement, and domain adaptation in improving vehicle detection in low-visibility environments.

While the aforementioned methods primarily address challenges in autonomous driving scenarios using  data captured by ego cameras. However, their applications are limited to the localized environment surrounding autonomous vehicles.  Network-level traffic flow and incident monitoring is typically achieved by roadside  cameras operated by state or local transportation agencies.  These cameras are commonly mounted on roadside utility poles and capture a broader view for effective traffic monitoring and incident management.  Such objectives are critical for enhancing the operational efficiency and safety of entire road networks.  These roadside cameras play a critical role in enhancing situational awareness by detecting incidents and disseminating alerts (e.g. speed warnings or hazard notifications) to drivers, thereby improving road safety. However, these camera are frequently characterized by low-resolution images and poor video quality. This challenge becomes even more pronounced under adverse conditions, such as nighttime or inclement weather, significantly complicating vehicle detection in these environments. Both roadside and ego cameras face common challenges in nighttime vehicle detection. These include difficulties in distinguishing vehicles from other objects under low illumination, glare from headlights, and insufficient detail captured by standard imaging sensors. Such limitations are particularly acute in rural settings, where minimal or inconsistent lighting exacerbates detection difficulties. Addressing these challenges is critical to improving the effectiveness of both camera types in their respective operational contexts. In this paper, we primarily focus on the nighttime vehicle detection from roadside cameras in rural settings.

Several studies have been conducted to address the challenges of nighttime vehicle detection. Fu et al. \cite{fu2021let} proposed a framework to improve nighttime object detection accuracy using a StyleMix-based method that generates day-night image pairs for training and a Kernel Prediction Network (KPN) to enhance nighttime-to-daytime image translation. While this framework aims to adapt models trained on daytime images for nighttime detection, the data used in their study was captured from a top-down perspective, and the resulting augmented nighttime images fail to accurately represent real roadside conditions. Specifically, the images lack critical challenges such as low illumination, poor contrast, and the presence of glare from headlights and reflections, which are common in real-world scenarios. Similarly, Guo et al. \cite{guo2023improved} employed CycleGAN to generate nighttime traffic images from daytime data, integrating these images with a Dense Traffic Detection Network (DTDNet) to enhance detection accuracy and address the scarcity of nighttime annotations. Nevertheless, their dataset is limited in perspective, as it was collected using phone cameras from specific angles. Consequently, the approach does not adequately account for real-world challenges such as low illumination and headlight glare, reducing its effectiveness in more complex and realistic nighttime environments. 

The suboptimal performance of detection models in nighttime rural scenarios stems from several interconnected challenges. Nighttime environments are characterized by low illumination and poor contrast, which hinder models' ability to distinguish vehicles from the background and accurately delineate vehicle boundaries. Additionally, intense headlight glare and reflections often confuse models, as these bright spots can obscure objects of interest or be misinterpreted as vehicles. With these issues, nighttime images frequently suffer from noise, motion blur, and low resolution, resulting from reduced sensor performance in low-light conditions. This degradation in data quality further impacts model accuracy and reliability. Another significant limitation is the scarcity of large, diverse, and annotated nighttime datasets.  Most existing datasets are predominantly consist of daytime images, leading to an imbalance that prevents models from effectively generalizing to nighttime conditions. Furthermore, domain adaptation remains a critical challenge, as models trained on daytime images struggle to perform in nighttime environments due to the stark differences in visual features and environmental conditions. These challenges collectively underscore the need for innovative approaches to enhance nighttime data quality, increase dataset diversity, and improve model adaptability for rural nighttime detection scenarios.

Generative models, particularly Generative Adversarial Networks (GANs), have emerged as powerful tools for augmenting datasets in scenarios where high-quality real-world data is scarce or difficult to obtain. Traditional GAN-based approaches, however,  rely heavily on the availability of paired data from two distinct domains (e.g., daytime and nighttime images). Previous studies, such as those by Fu et al. \cite{fu2021let} and Guo et al. \cite{guo2023improved}, utilized CycleGAN for day-to-night image style transfer, generating synthetic nighttime images to enhance dataset diversity and improve the performance of detection models. While these methods can effectively achieve day-to-night transfer, they often fail to address more complex challenges, such as accurately replicating headlight effects, which is critical for vehicle detection in rural nighttime settings where roadside lighting is absent. Recently, the rapid advancement of Generative Pretrained Transformers (GPTs) has led to the development of various AI tools capable of performing image style transfer tasks, including text-based image editing.  Motivated by these advancements, we explored text-based image editing tools leveraging diffusion models \cite{kawar2023imagic}. Using a prompt such as "Given the daytime image, transfer it into a nighttime setting without ambient light and turn on the headlights of all the vehicles," we observed promising results for day-to-night style transfer. However the generated images exhibited poor and unrealistic headlight modeling, highlighting the limitations of GPT-like models in accurately simulating rural nighttime transportation conditions.

A key limitation of existing approaches lies in their inability to accurately model headlight effects, as the distribution of headlight illumination is governed by complex physical principles that are challenging to replicate using  simple domain mapping techniques. At night, headlights serve as the most prominent and reliable vehicle feature for detection. Several researchers have explored nighttime vehicle detection through headlight detection, tracking, and pairing methods \cite{zou2015robust, parvin2021vision}.  
While these techniques perform well in low-light scenarios, they are impractical for roadside camera settings in rural areas. To address these challenges, we propose a novel framework that (1) enables the augmentation of annotated nighttime images directly from daytime images, and (2) realistically model headlight effects using the CARLA simulator for image style transfer. To the best of our knowledge, this is the first study to leverage CARLA-generated synthetic data for both day-to-night image style transfer and headlight effects modeling.  Our framework offers a novel and effective solution to enhance vehicle detection in challenging rural nighttime environments.

\section{Framework}

\begin{figure}[htbp!]
    \centering
    \includegraphics[width=1\linewidth]{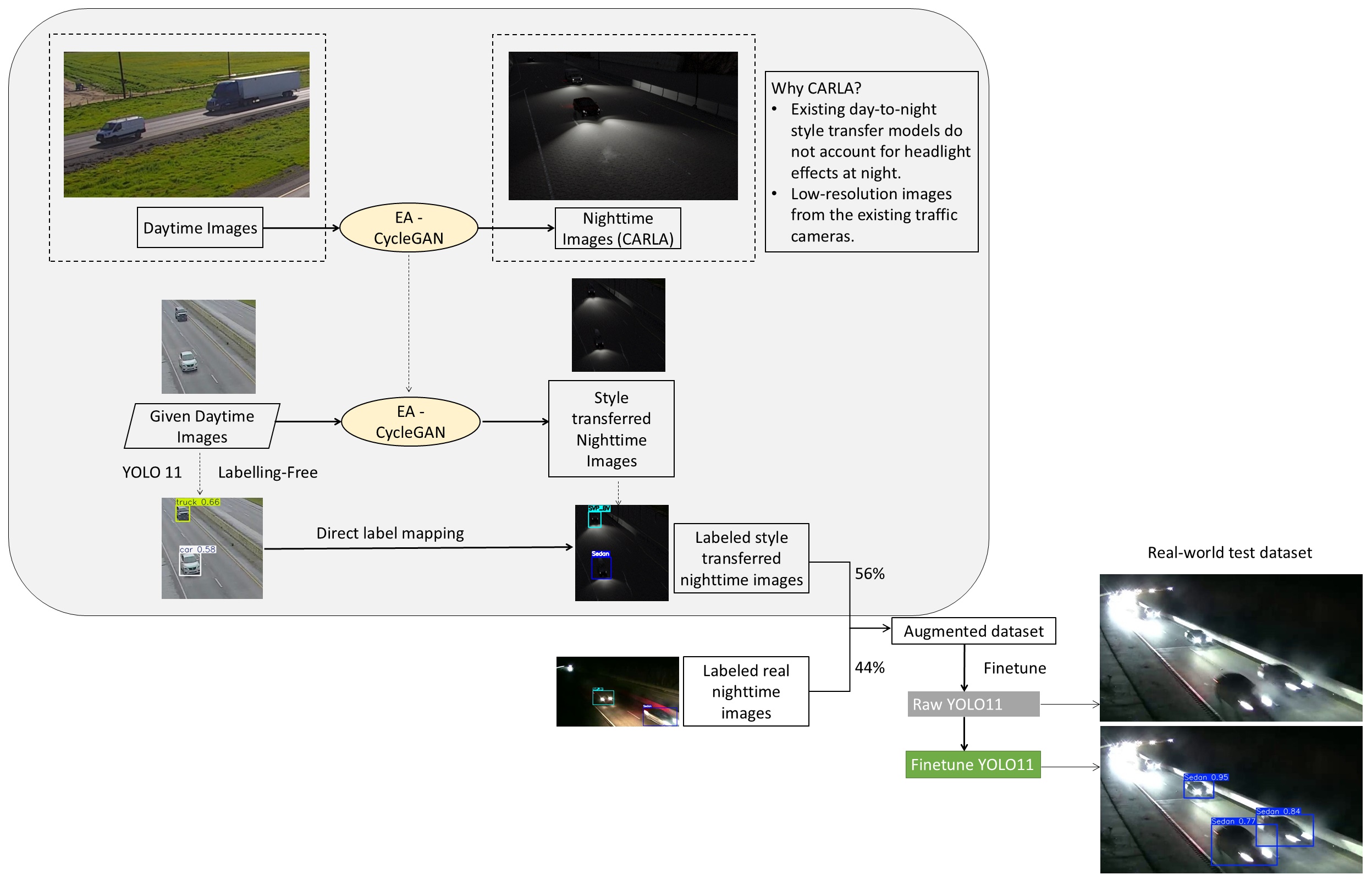}
    \caption{Framework overview.}
    \label{fig:framework}
\end{figure}

Our proposed framework, illustrated in Figure \ref{fig:framework}, introduces a novel labeling-free data augmentation method that  enables realistic day-to-night image style transfer using synthetic data generated by CARLA \cite{dosovitskiy2017carla}.  The framework comprises two main components: 

\begin{enumerate}
	
\item Synthetic nighttime data generation under rural settings: This component leverages the CARLA simulator to generate synthetic nighttime images that incorporates realistic headlight effects and varying illumination conditions, as observed from roadside cameras in rural environments.  The CARLA simulator is integral to this process, as it can faithfully model vehicle headlight effects at night, effectively addressing the limitations of existing AI models that often fail to capture headlight effects during day-to-night style transfers. 

\item Day-to-night style transfer process: To address data scarcity of nighttime road scene images in rural environments, a CycleGAN model is trained to perform day-to-night style transfer. 
Daytime images are collected and processed using the state-of-the-art YOLO11 model \cite{khanam2024yolov11} to perform vehicle detection and classification. The resulting annotations are directly mapped to the style-transferred nighttime images, enabling the creation of an augmented nighttime dataset without additional labeling effort. To enhance dataset diversity and realism, the final augmented dataset combines human-labeled real nighttime low-light images (44\%) with style-transferred images (56\%). This dataset is subsequently used to fine-tune the YOLO11 model, which is evaluated against its raw counterpart on a real-world nighttime test dataset. 

\end{enumerate}

By combining realistic synthetic data generation with effective style transfer techniques and automated annotation mapping, our framework addresses critical challenges in rural nighttime vehicle detection, offering a novel and practical solution to improve model performance in real-world scenarios.

\section{Method}

This section introduces our proposed method, which addresses the challenges of nighttime vehicle detection in rural environments through three key steps: (1) Synthetic Nighttime Data Generation: The process of generating realistic nighttime images is described, where the CARLA simulator is utilized to incorporate critical features such as headlight effects and varying illumination conditions.
(2) Day-to-Night Image Style Transfer: The model architecture employed for performing day-to-night image translation is presented, enabling the creation of nighttime images that closely resemble real-world scenarios.
(3) Labeling-Free Data Augmentation for Nighttime Images: The approach for achieving labeling-free augmentation is described, where annotations from daytime images are directly mapped onto style-transferred nighttime images, facilitating the development of a robust augmented dataset.

\subsection{Synthetic nighttime data generation}

As discussed in the previous section, the primary challenges in improving nighttime vehicle detection arise from the low quality of roadside camera images and the difficulty of collecting sufficiently large and diverse datasets. To address these issues, synthetic nighttime images are generated using CARLA  \cite{dosovitskiy2017carla}, a widely used open-source platform primarily designed for autonomous driving research. CARLA offers extensive control over various environmental and operational parameters, such as weather conditions, lighting, vehicle types, headlight settings (e.g., low-beam, high-beam), as well as camera positions and viewing angles. These customizable options enable the creation of a comprehensive and diverse dataset that accurately reflect real-world rural transportation settings. In particular, for rural highway safety research, the simulator allows for the strategic placement of cameras in critical locations, such as curves and ramps, where lower speed limits are often imposed \cite{malik2022carla}.

\begin{figure}[htbp!]
    \centering
    \includegraphics[width=0.8\linewidth]{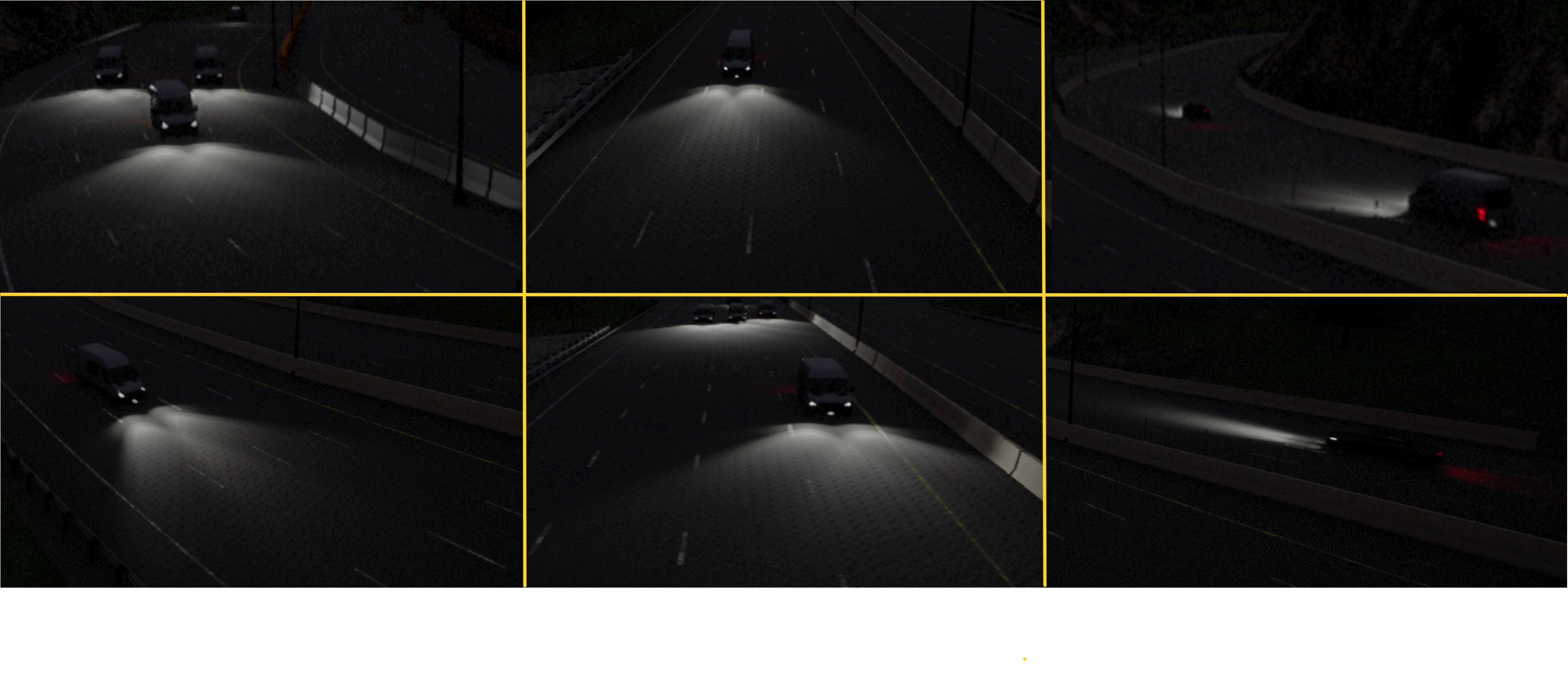}
    \caption{CARLA examples (From left to right, first column: side-view approaching; second column: center-view approaching; third column: side-view departing).}
    \label{fig:CARLA samples}
\end{figure}

To closely mimic realistic rural environments, synthetic images were collected under the following scenarios: (1) departing and approaching vehicles relative to cameras, (2) side-view and top-view perspectives, and (3) scenes with multiple vehicles and single vehicles. Several representative examples are presented in Figure \ref{fig:CARLA samples}. It is important to note that, in this study, all synthetic images were generated under clear weather conditions with no environmental modifications.

\subsection{Day-to-night image style transfer}

The Efficient Attention GAN (EAGAN) \cite{efficient_attention_GAN} builds upon the CycleGAN framework by integrating efficient attention blocks into the generator networks while enabling attention sharing between corresponding encoder and decoder blocks.  This mechanism allows the re-utilization of the long-range dependencies computed from the source-domain images during the reconstruction of their target-domain counterparts. This design makes EAGAN a robust choice for high-quality image-to-image (I2I) translation tasks, particularly in scenarios where maintaining consistency between domains is critical.

In this study, the EAGAN architecture is adopted to perform day-to-night style transfer in rural environments. The model is trained using datasets from two specific domains: real-world daytime images and CARLA virtual nighttime images.

The I2I translation task generally considers transforming image \( x \) from domain \( X \) (daytime) to image \( y \) in domain \( Y  \) (nighttime), represented as mappings:  \( G: x \to y, \ F: y \to x \), where \( G \) and \( F \) are generator networks. The objective is to ensure that the distributions \( G(X) \) and \( F(Y) \) are indistinguishable from \( X \) and \( Y \), respectively, while preserving semantic information and cycle consistency.

\begin{figure}[htbp]
    \centering
    \includegraphics[width=0.7\linewidth]{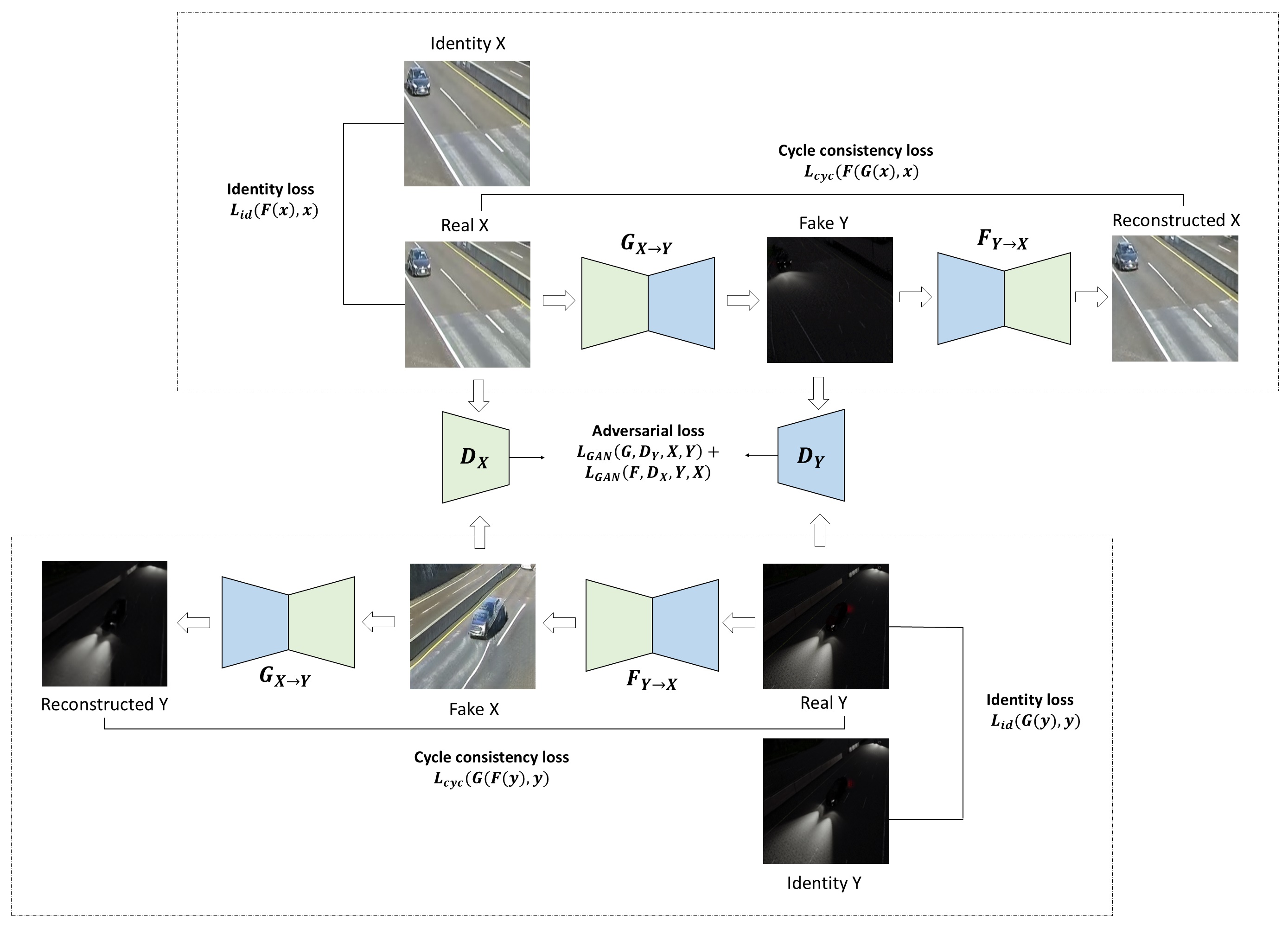}
    \caption{The information flow of EAGAN.}
    \label{fig:information flow ECGAN}
\end{figure}

To train the EAGAN for the data augmentation purpose from daytime images and CARLA nighttime images, the model input consists of \{Real X, Real Y\}, where Real X, and Real Y are from the domains X and Y. The detailed information flow is shown in Figure \ref{fig:information flow ECGAN}. Following the standard training process for GANs, the discriminators and generators are trained simultaneously by optimizing a min-max adversarial objectives. Instead of the traditional adversarial loss, the least-square adversarial loss proposed by \cite{mao2017least} was used due to improved stability, which also encourages the generator to produce realistic images indistinguishable by the discriminator:

\begin{equation}
\min_G L_{\text{GAN}}(G) = \mathbb{E}_{x \sim p_{\text{data}}(x)} \left[ \left( D(G(x)) - 1 \right)^2 \right]
\end{equation}

Where \( p_{\text{data}}(x) \) denotes the true data distribution of domain \( X \), comprising real daytime images. \( D(G(x)) \) represents the output of the discriminator, which assigns a score between 0 and 1. Ideally, \( D(G(x)) \) should approach 1 if the generated nighttime image appears realistic. The term \((D(G(x)) - 1)^2\) imposes a penalty on the generator if the discriminator fails to classify \( G(x) \) as realistic, i.e., when the score deviates from 1.

\begin{equation}
\min_D L_{\text{GAN}}(D) = \mathbb{E}_{y \sim p_{\text{data}}(y)} \left[ \left( D(y) - 1 \right)^2 \right] + \mathbb{E}_{x \sim p_{\text{data}}(x)} \left[ D(G(x))^2 \right]
\end{equation}

Where \( p_{\text{data}}(y) \) denotes the true data distribution of domain \( Y \), i.e., CARLA nighttime images. The term \( \mathbb{E}_{y \sim p_{\text{data}}(y)} \left[ \left( D(y) - 1 \right)^2 \right] \) ensures that the discriminator assigns a score close to 1 to authentic CARLA nighttime images from domain \( Y \).  Conversely, \( \mathbb{E}_{x \sim p_{\text{data}}(x)} \left[ D(G(x))^2 \right] \) penalizes the discriminator if it assigns a high score to the nighttime images \( G(x) \) generated from real daytime images \( x \).  This dual mechanism maintains the discriminator's reliability in distinguishing between real and generated samples.

In addition to the least square adversarial loss, cycle consistency is enforced between the two generators to ensure the reversibility of the translation process. Specifically, when an image is passed sequentially through both generators, the reconstructed image should closely resemble the original. To achieve this, a cycle consistency loss term is added to the objective function alongside the adversarial loss:

\begin{equation}
L_{\text{cyc}}(G, F) = \mathbb{E}_{x \sim P_X} \left[ \lVert F(G(x)) - x \rVert_1 \right] + \mathbb{E}_{y \sim P_Y} \left[ \lVert G(F(y)) - y \rVert_1 \right].
\end{equation}

Furthermore, an identity loss term, as proposed by  \cite{taigman2016unsupervised}, is included as well. By leveraging Identity mappings for domains X and Y, the generators are encouraged to make minimal alterations to the input images \( x \) and \( y \) when they already belongs to the target domain. This constraint helps the generators to better preserve the original tint and coloration of the input images. The identity loss is expressed as:

\begin{equation}
L_{\text{id}}(G, F) = \mathbb{E}_{y \sim P_Y} \left[ \lVert G(y) - y \rVert_1 \right] + \mathbb{E}_{x \sim P_X} \left[ \lVert F(x) - x \rVert_1 \right].
\end{equation}
As a result, the overall training objective for EAGAN combines adversarial loss, cycle consistency loss, and identity loss, and is written as:

\begin{equation}
L(G, F, D_X, D_Y) = L_{\text{GAN}}(G, D_Y, X, Y) + L_{\text{GAN}}(F, D_X, Y, X) + \lambda_{\text{cyc}} L_{\text{cyc}}(G, F) + \lambda_{\text{id}} L_{\text{id}}(G, F),
\end{equation}

where \(\lambda_{\text{cyc}}\) and \(\lambda_{\text{id}}\) are weighting parameters for the respective loss terms.

\subsection{Labeling-free data augmentation for nighttime images}

The You Only Look Once (YOLO) family of models has revolutionized object detection, offering real-time detection and high accuracy. YOLO11, the latest version \cite{khanam2024yolov11}, builds upon this legacy with attention mechanisms, deeper feature extraction layers, and an anchor-free detection approach. It is specifically designed to address challenges such as detecting small, occluded, or fast-moving vehicles. By integrating strengths of CNNs and self-attention mechanism, YOLO11 improves both detection accuracy and computational efficiency, making it well-suited for real-world applications. In our study, YOLO11 is applied as an "annotator" to automatically annotate daytime images.  The obtained labels from daytime images are directly applied to the style-transferred nighttime images as the objects of interest (i.e., vehicles) will remain in same locations.  This allows us to leverage the accurate vehicle detection capability of the YOLO11 model to automatically obtain labels for its style-transferred nighttime counterparts.

The original YOLO11 model was pretrained on the COCO dataset \cite{lin2014microsoft}, which includes vehicle classes of car, bus, and truck that are relevant to rural settings.  For this study, our focus is on classifying  two specific vehicle categories: class 0 (Sedan) and class 1 (SVP-BV).  The SVP-BV category includes SUV, Van, Pick-UP, and Bigger Vehicles.  To align with the new vehicle categories, the COCO vehicle classes are remapped as follows: (1) cars -> Sedan; (2) bus and truck -> SVP-BV.

The augmented dataset is obtained by assembling the style-transferred nighttime images generated by the EAGAN model with the corresponding labels predicted by YOLO11.

\section{Experiments}

\subsection{Data}
Our data was gathered from multiple public traffic cameras in California, which includes both daytime and nighttime images, serving distinct purposes for training and testing. Specifically, the data is organized into three categories: (1) training datasets for the EAGAN model, (2) fine-tuning datasets for the YOLO11 model, and (3) an evaluation dataset for comparing the performance of the original YOLO11 model and the fine-tuned version.

A detailed summary of each dataset is provided in Table~\ref{tab:dataset_details}.

\begin{table}[h!]
\centering
\caption{Dataset details.}
\label{tab:dataset_details}
\begin{tabular}{|l|l|l|l|}
\hline
\textbf{Dataset} & \textbf{Split}      & \textbf{Description}                                   & \textbf{Number of Images} \\ \hline
\multirow{4}{*}{EAGAN Training} 
                 & Train A             & Real daytime images                                    & 239                       \\ \cline{2-4} 
                 & Train B             & CARLA images                                           & 413                       \\ \cline{2-4} 
                 & Test A              & Real daytime images                                    & 82                        \\ \cline{2-4} 
                 & Test B              & CARLA images                                           & 50                        \\ \hline
\multirow{3}{*}{YOLO11 Fine-Tuning}   
                 & Train               & 124 real nighttime images + 163 augmented images      & 287                       \\ \cline{2-4} 
                 & Validation          & 43 real nighttime images + 20 augmented images        & 63                        \\ \cline{2-4} 
                 & Test                & 20 real nighttime images + 10 augmented images        & 30                        \\ \hline
Evaluation& Test                & Real nighttime images                                  & 38                        \\ \hline
\end{tabular}
\end{table}

\subsection{Image style transfer}
For the image style transfer with EAGAN, we targeted two domains: domain X, which consists of daytime images in real-world settings, and domain Y, comprising CARLA-generated nighttime images.  The EAGAN was trained for 200 epochs with scheduled learning rate (Equation 6) that was initialized at 0.0002 and started to decrease linearly after the 100th epoch.  This learning rate decay strategy ensure smooth model convergence, leading to better generalization and performance \cite{you2019does}.

\begin{equation}
\text{lr}(t) =
\begin{cases} 
\text{lr}_0 & \text{if } t \leq n_{\text{epochs}} \\
\text{lr}_0 \cdot \left( 1 - \frac{t - n_{\text{epochs}}}{n_{\text{epochs\_decay}}} \right) & \text{if } n_{\text{epochs}} < t \leq n_{\text{epochs}} + n_{\text{epochs\_decay}} \\
0 & \text{if } t > n_{\text{epochs}} + n_{\text{epochs\_decay}}
\end{cases}
\end{equation}

Where
\( \text{lr}_0  = \)  initial learning rate 
\( t = \)  current epoch
\( n_{\text{epochs}} = \)  the number of epochs that the learning rate decay starts 
\( n_{\text{epochs\_decay}} =\) the number of epochs that the learning rate decay ends.

Table 2 shows the detailed parameter settings for EAGAN training. 

\begin{table}[ht]
\centering
\caption{Training Parameter Settings for EAGAN.}
\label{table:essential-parameters}
\begin{tabular}{|l|l|l|}
\hline
\textbf{Parameter}          & \textbf{Value} & \textbf{Description} \\ \hline
\texttt{$n_epochs$}        & 100                 & Epochs with constant learning rate \\ \hline
\texttt{$n_epochs\_decay$} & 200& Epochs for linear learning rate decay \\ \hline
\texttt{$lr_0$}               & 0.0002              & Initial learning rate \\ \hline
\texttt{$\beta_1$}            & 0.5                 & Momentum term of Adam optimizer \\ \hline
\texttt{input\_size}      & $256 \times 256$       & Size of input images \\ \hline
\texttt{$\lambda_{\text{cyc}}$}        & 10.0     & Weight for cycle consistency loss \\ \hline
\texttt{$\lambda_{\text{id}}$} & 0.5                 & Weight for identity loss \\ \hline
\end{tabular}
\end{table}

\begin{figure}[H]
    \centering
    \includegraphics[width=0.9\linewidth]{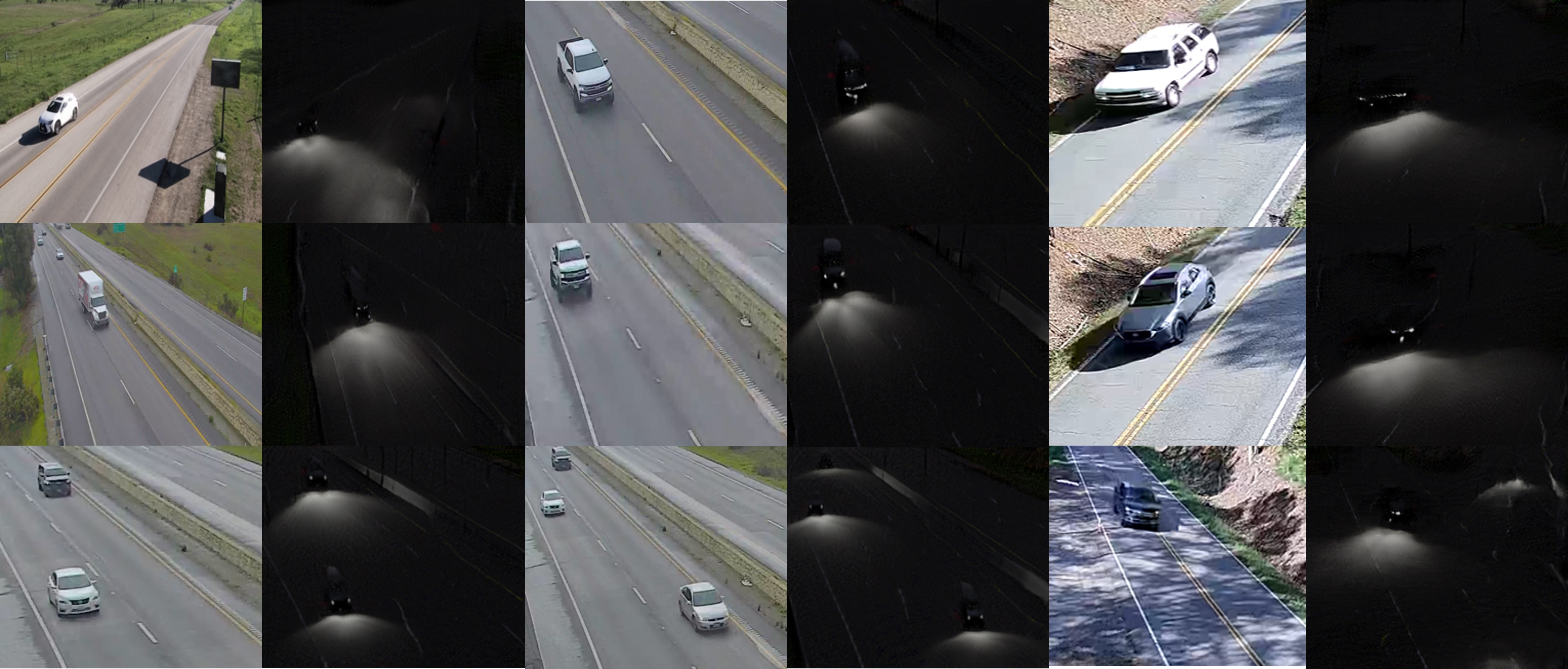}
    \caption{Test examples of the trained EAGAN (From left to right:the first, third, and fifth columns are original daytime images; the second, fourth, and sixth columns are the corresponding style-transferred nighttime images, respectively).}
    \label{fig:Test examples of well-trained EAGAN}
\end{figure}

Each epoch takes approximately 150 seconds, with the entire training process completing in about 8 hours on a single NVIDIA A6000 GPU. 
Figure~\ref{fig:Test examples of well-trained EAGAN} showcases test examples from our trained EAGAN model. The results confirm successful day-to-night translation, including effective addition of headlight features. Notably, the model accurately places headlights in the correct locations of the vehicles, demonstrating its ability to reliably locating the vehicles and identifying their positions. Interestingly, some shadow-related effects were observed: 

\begin{enumerate}
\item Vehicle Shadows under Sunlight: For shadows cast by vehicles in sunny conditions (e.g., rows 1 and 2 in columns 5 and 6 of Figure 4), the model tends to interpret the shadow in front of the car's bumper as part of the vehicle. This results in a slight angular misalignment between the illuminated headlights and the front of the car in the transferred image. However, this minor deviation does not impair the model's ability to recognize vehicles at night. 
\item Vehicles in Shadowed Areas beneath Trees: When vehicles are passing tree-shaded areas (e.g., row 3 in columns 5 and 6 of Figure 4), the blending of vehicle features with the blotchy tree shadows create challenges for the model. These shadowed regions act as noise, degrading the quality of the transferred images and negatively impacting downstream tasks.
\end{enumerate}

While these shadow effects introduce some artifacts, the overall performance of the EAGAN model remains robust in generating high-quality day-to-night image translation.

\subsection{Nighttime vehicle detection and classification}
\begin{figure}[H]
    \centering
    \includegraphics[width=0.9\linewidth]{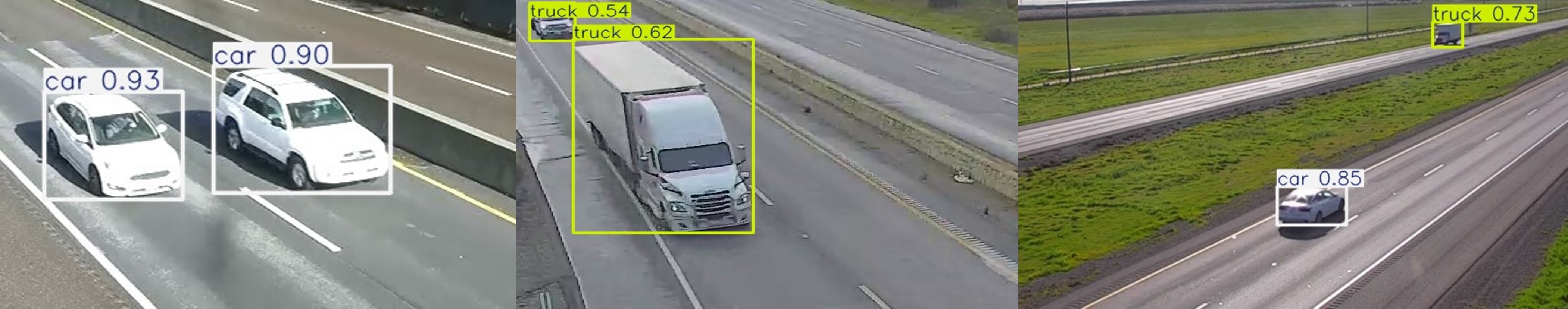}
    \caption{Original YOLO11-Small model predictions for auto-labeling.}
    \label{fig:Raw YOLO11-Small model}
\end{figure}

For this experiment, the YOLO11-Small model was employed. Figure~\ref{fig:Raw YOLO11-Small model} showcase sample predictions generated by the original YOLO11-Small model, which serve as labels for their style-transferred nighttime images.
\vspace{-1em}

\vspace{-1em}

Although CARLA can generate realistic nighttime road scene images, there are still subtle differences in appearance compared to real-world nighttime road scenes. To address this domain adaption gap, we incorporated a selection of manually annotated real-world nighttime images into the training dataset. This approach allows the model to learn relevant features from both CARLA-generated and
real-world nighttime images, enhancing its overall performance and robustness. 

To fine-tune the model, a learning rate scheduling strategy was implemented for different components of the model. Initailly, the backbone was fine-tuned with a learning rate of 0.0001 for 50 epochs. Subsequently, the backbone network was frozen, and a learning rate of 0.00005 was applied exclusively to the neck network for another 50 epochs. Finally, both the backbone and neck networks were frozen, and a learning rate of 0.00001 was applied to the head network for an additional 50 epochs. This block-wise adaptation strategy, distinct from the approach used in EAGAN training, facilitates enhanced convergence and improved generalization.

\begin{figure}[htbp!]
    \centering
    \includegraphics[width=0.8\linewidth]{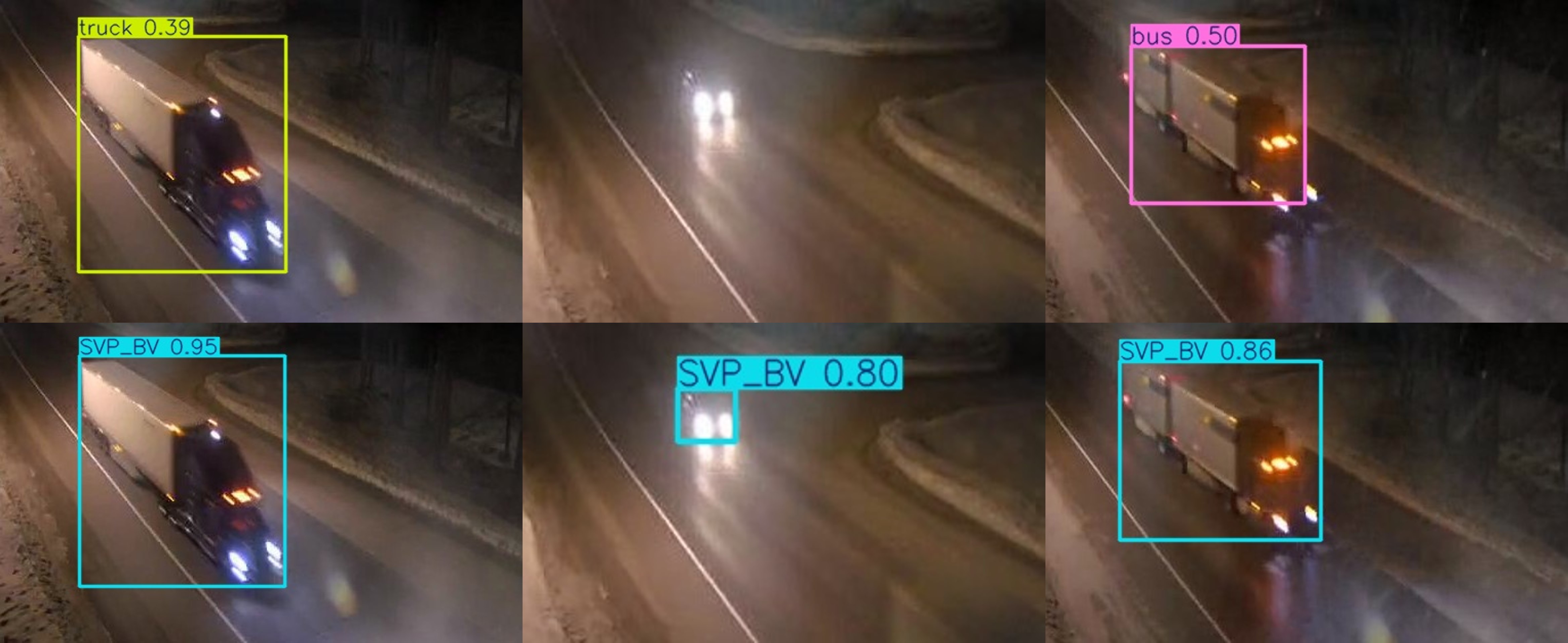}
    \caption{Comparison of predictions on single-vehicle images with ambient light (Top row: Original YOLO11 model; Bottom row: fine-tuned YOLO11 model).}
    \label{fig:Comparison of the predictions (1) }
\end{figure}

\begin{figure}[htbp!]
    \centering
    \includegraphics[width=0.8\linewidth]{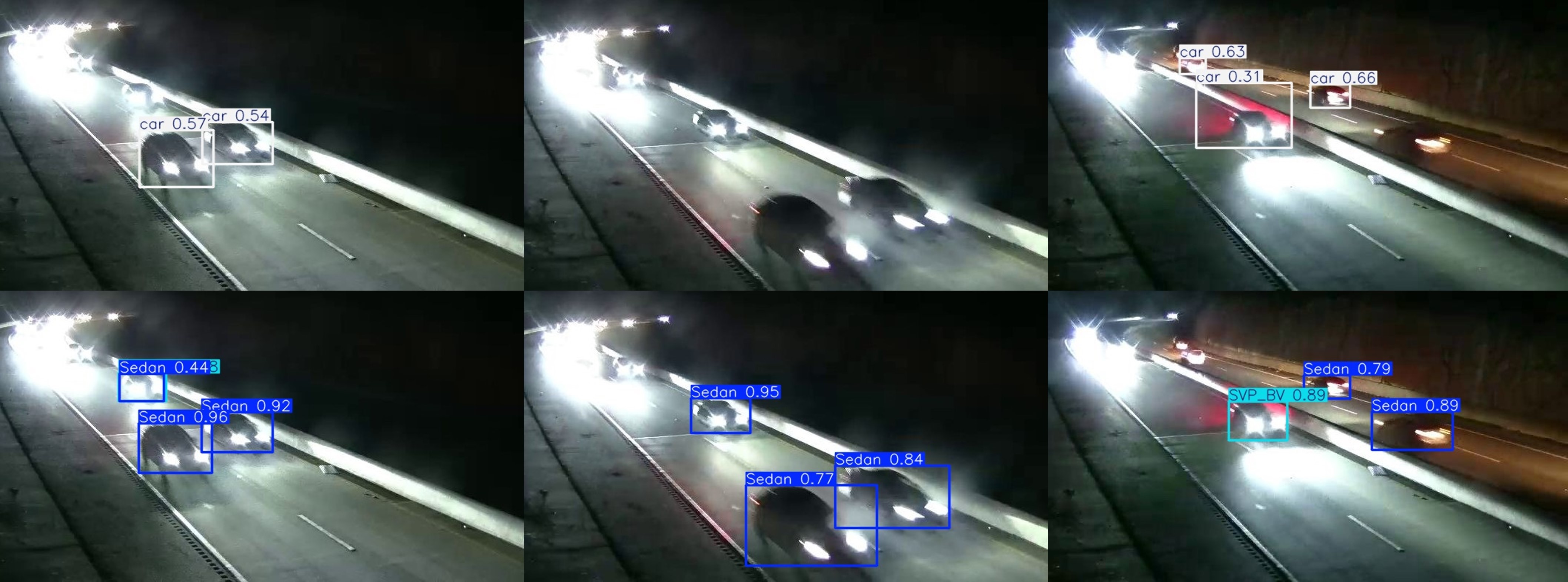}
    \caption{Comparison of predictions on multiple-vehicle images without ambient light (Top row: Original YOLO11 model; Bottom row: fine-tuned YOLO11 model).}
    \label{fig:Comparison of the predictions (2)}
\end{figure}

\begin{figure}[htbp!]
    \centering
    \includegraphics[width=0.8\linewidth]{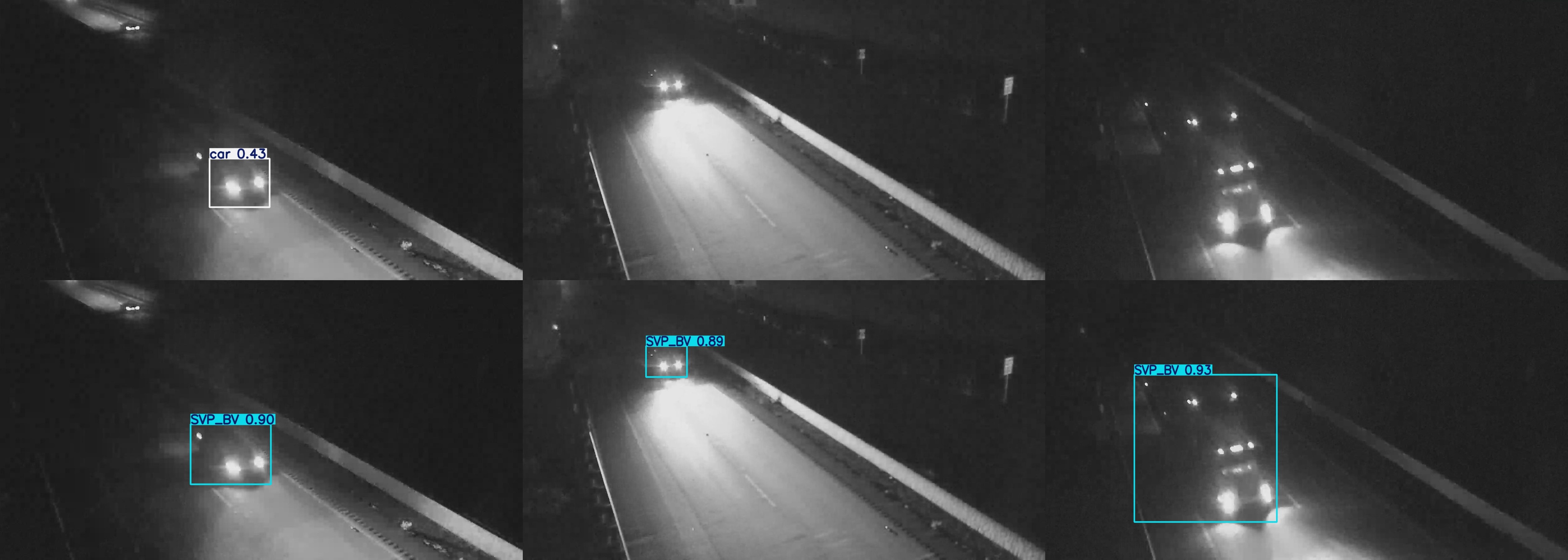}
    \caption{Comparison of predictions on gray-scale images without ambient light (Top row: Original YOLO11 model; Bottom row: fine-tuned YOLO11 model).}
    \label{fig:Comparison of the predictions (3) }
\end{figure}

For evaluation, a selection of representative real-world nighttime images captured by roadside traffic cameras in rural areas was analyzed, as shown in Figures~\ref{fig:Comparison of the predictions (1) }–\ref{fig:Comparison of the predictions (3) }. These nighttime images present various challenges, including low ambient light, poor image quality, and issues caused by headlight glare. The original YOLO11 model frequently struggled to distinguish vehicles from the background, even under relatively favorable lighting conditions, and often produced low confidence scores when vehicles were detected. In contrast, the fine-tuned YOLO11 model, trained on the augmented dataset, achieved a 100\% detection success rate, with significantly higher confidence scores, demonstrating the effectiveness of the proposed framework.

Table~\ref{tab:model_performance} presents detailed classification results across various metrics for the original and fine-tuned YOLO11 models:

\begin{table}[htbp!]
\centering
\caption{Performance comparison of the original and fine-tuned YOLO11 models.}
\label{tab:model_performance}
\begin{tabular}{|l|l|c|c|c|c|}
\hline
\textbf{Model}        & \textbf{Class} & \textbf{Bounding Box Precision} & \textbf{Recall}& \textbf{mAP50} & \textbf{mAP50-95} \\ \hline
\multirow{3}{*}{Raw YOLO11} 
                      & all            & 0.559           & 0.246        & 0.259          & 0.160            \\ \cline{2-6} 
                      & car            & 0.206           & 0.385        & 0.203          & 0.120            \\ \cline{2-6} 
                      & truck          & 0.912           & 0.108        & 0.316          & 0.200            \\ \hline
\multirow{3}{*}{Fine-tuned YOLO11} 
                      & all            & 0.631           & 0.883        & 0.758          & 0.559            \\ \cline{2-6} 
                      & Sedan          & 0.514           & 0.846        & 0.592          & 0.397            \\ \cline{2-6} 
                      & SVP\_BV        & 0.748           & 0.919        & 0.925          & 0.721            \\ \hline
\end{tabular}
\end{table}

Table~\ref{tab:model_performance} reveals significant improvements across classes, indicating that the fine-tuned model effectively captures most vehicles in the nighttime scenes, addressing the key limitation of current state-of-the-art object detection models. The Consistent gains in mAP metrics further highlight the fine-tuned model’s robustness in detecting and localizing vehicles under challenging nighttime conditions. Class-specific refinements enhance detection for both smaller vehicles (Sedan) and larger ones (SVP-BV). Notably, the Fine-tuned model shows a slightly lower bounding box precision for the SVP-BV class, largely due to the diverse mix of vehicle types in this newly defined class.

\section{Conclusion}

In this work, we proposed a novel framework for enhancing nighttime vehicle detection, featuring a labeling-free method to create an augmented dataset to fine-tuning object detection models with improve performance for nighttime conditions. We employed the EAGAN as the image translator to generate corresponding nighttime images from their daytime counterparts. Additionally, we adopted different learning rate scheduling strategies during EAGAN training and YOLO11 fine-tuning to ensure smooth convergence and enhanced generalization. A performance comparison between the original YOLO11 model and the fine-tuned version demonstrated that the YOLO11 model fine-tuned with the augmented dataset significantly outperformed the original YOLO11 model for nighttime vehicle detection. Its ability to detect and localize the vehicles with high confidence highlights the effectiveness of fine-tuning with properly augmented data, making it a more reliable solution for real-world applications. 

\vspace{1em}
Nevertheless, we acknowledge several limitations that should be addressed in future research: 
(1) While CARLA currently offers a wide variety of vehicle types, it still lacks coverage of all vehicle types on the road, particularly the tractor-trailers and RVs, which limits the diversity of synthetic data. Additionally, the headlights in CARLA need further refinement to better replicate glare effects observed in real-world settings. (2) Although the EAGAN model incorporates an attention-sharing mechanism in the generators of CycleGAN, future research could explore alternative mechanisms to more effectively address the observed shadow effects. (3) For proof of the concept, the training and testing datasets utilized in this study are relatively small. Future work should consider significantly expanding the datasets using our proposed data augmentation approach, which is expected to further enhance model performance and robustness.

\section*{Acknowledgment}
This work was supported by the U.S. Department of Transportation (USDOT) University Transportation Center (UTC) Program under Grant 69A3552348304.

\bibliographystyle{unsrt} 
\bibliography{references}  
\end{document}